\documentclass[12pt,letterpaper]{article}
\usepackage[letterpaper,vmargin={1in},hmargin={1in}]{geometry}
\usepackage{etex}
\usepackage{multicol}
\usepackage{amsthm}
\usepackage[bookmarks=true]{hyperref}

\usepackage{comment}
\usepackage[binary-units]{siunitx}
\usepackage{relsize}
\usepackage{ifthen}
\usepackage[colorinlistoftodos]{todonotes}

\usepackage[vlined,ruled,linesnumbered]{algorithm2e}
\usepackage{graphics} % for pdf, bitmapped graphics files
\usepackage{rotating}
\usepackage{color}
\usepackage{enumerate}
\usepackage[T1]{fontenc}
\usepackage{psfrag}
\usepackage{epsfig} % for postscript graphics files
\usepackage{booktabs}
\usepackage{graphicx,url}
\usepackage{multirow}
\usepackage{array}
\usepackage{latexsym}
\usepackage{amsfonts}
\usepackage{amsmath}
\usepackage{amssymb}
\usepackage{xstring}
\usepackage[noend]{algorithmic}
\usepackage{multirow}
\usepackage{xcolor}
\usepackage{prettyref}
\usepackage{flexisym}
\usepackage{bigdelim}
\usepackage{breqn} % load this last
\usepackage{listings}
\let\labelindent\relax
\usepackage{enumitem}
\usepackage{xspace}
\usepackage{bm}
\graphicspath{{./figures/}}
\usepackage{tikz}
\usetikzlibrary{matrix,calc}

\usepackage{mdwlist}

\let\itemize\undefined
\makecompactlist{itemize}{stditemize}

\usepackage{amsfonts,amssymb,amsmath}
\usepackage{thmtools} 
\usepackage{thm-restate}

\newtheoremstyle{mystyle}%                % Name
  {}%                                     % Space above
  {}%                                     % Space below
  {\itshape}%                                     % Body font
  {}%                                     % Indent amount
  {\bfseries}%                            % Theorem head font
  {.}%                                    % Punctuation after theorem head
  { }%                                    % Space after theorem head, ' ', or \newline
  {\thmname{#1}\thmnumber{ #2}\thmnote{ (#3)}}% % Theorem head spec (can be left empty, meaning `normal')
\theoremstyle{mystyle}
\newrefformat{prob}{Problem\,\ref{#1}}
\newrefformat{def}{Definition\,\ref{#1}}
\newrefformat{sec}{Section\,\ref{#1}}
\newrefformat{sub}{Section\,\ref{#1}}
\newrefformat{prop}{Proposition\,\ref{#1}}
\newrefformat{app}{Appendix\,\ref{#1}}
\newrefformat{alg}{Algorithm\,\ref{#1}}
\newrefformat{cor}{Corollary\,\ref{#1}}
\newrefformat{thm}{Theorem\,\ref{#1}}
\newrefformat{lem}{Lemma\,\ref{#1}}
\newrefformat{fig}{Fig.\,\ref{#1}}
\newrefformat{tab}{Table\,\ref{#1}}

\newcommand{\bdmath}{\begin{dmath}}
\newcommand{\edmath}{\end{dmath}}
\newcommand{\beq}{\begin{equation}}
\newcommand{\eeq}{\end{equation}}
\newcommand{\bdm}{\begin{displaymath}}
\newcommand{\edm}{\end{displaymath}}
\newcommand{\bea}{\begin{eqnarray}}
\newcommand{\eea}{\end{eqnarray}}
\newcommand{\beal}{\beq \begin{array}{ll}}
\newcommand{\eeal}{\end{array} \eeq}
\newcommand{\beas}{\begin{eqnarray*}}
\newcommand{\eeas}{\end{eqnarray*}}
\newcommand{\ba}{\begin{array}}
\newcommand{\ea}{\end{array}}
\newcommand{\bit}{\begin{itemize}}
\newcommand{\eit}{\end{itemize}}
\newcommand{\ben}{\begin{enumerate}}
\newcommand{\een}{\end{enumerate}}

\newcommand{\hide}[1]{}

\newcommand{\hiddenText}{{\color{gray} hidden text.}}
\newcommand{\hideWithText}[1]{\hiddenText}

\newcommand{\blue}[1]{{\color{blue}#1}}

\newcommand{\linkToPdf}[1]{\href{#1}{\blue{(pdf)}}}
\newcommand{\linkToPpt}[1]{\href{#1}{\blue{(ppt)}}}
\newcommand{\linkToCode}[1]{\href{#1}{\blue{(code)}}}
\newcommand{\linkToWeb}[1]{\href{#1}{\blue{(web)}}}
\newcommand{\linkToVideo}[1]{\href{#1}{\blue{(video)}}}
\newcommand{\linkToMedia}[1]{\href{#1}{\blue{(media)}}}
\newcommand{\award}[1]{\xspace} % {{\red{#1}}} % omit awards

\usepackage{float}
\usepackage{graphicx}
\usepackage{epstopdf}
\usepackage[font=small]{caption}
\usepackage{epsfig}
\usepackage{cleveref}
\usepackage{subcaption}
\theoremstyle{definition}
\usepackage{xr}

\title{Learning Correspondence for Deformable Objects}
\date{}
\author{%
  Priya Sundaresan, Aditya Ganapathi, Harry Zhang, Shivin Devgon
}

\begin{document}

\maketitle
\begin{figure}[h!]%
    \centering
    \subfloat[Rope (sphere added for asymmetry)]{{\includegraphics[width=4cm]{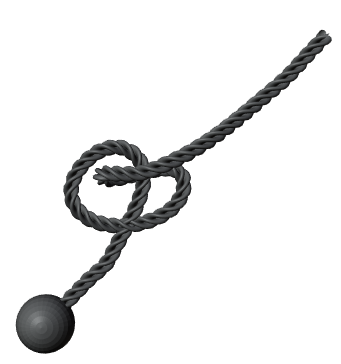} }}%
    \qquad
    \subfloat[Cloth]{{\includegraphics[width=4cm]{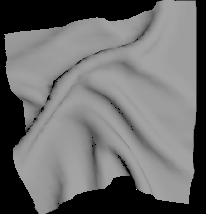} }}%
    \caption{We investigate finding correspondences across images of simulated rope and cloth. }%
    \label{fig:example}%
\end{figure}
\begin{abstract}

We investigate the problem of pixelwise correspondence for deformable objects, namely cloth and rope, by comparing both classical and learning-based methods. We choose cloth and rope because they are traditionally some of the most difficult deformable objects to analytically model with their large configuration space, and they are meaningful in the context of robotic tasks like cloth folding, rope knot-tying, T-shirt folding, curtain closing, etc. The correspondence problem is heavily motivated in robotics, with wide-ranging applications including semantic grasping, object tracking, and manipulation policies built on top of correspondences. We present an exhaustive survey of existing classical methods for doing correspondence via feature-matching, including SIFT, SURF, and ORB, and two recently published learning-based methods including TimeCycle and Dense Object Nets. We make three main contributions: (1) a framework for simulating and rendering synthetic images of deformable objects, with qualitative results demonstrating transfer between our simulated and real domains (2) a new learning-based correspondence method extending Dense Object Nets, and (3) a standardized comparison across state-of-the-art correspondence methods. Our proposed method provides a flexible, general formulation for learning temporally and spatially continuous correspondences for nonrigid (and rigid) objects. We report root mean squared error statistics for all methods and find that Dense Object Nets outperforms baseline classical methods for correspondence, and our proposed extension of Dense Object Nets performs similarly. The link to our presentation, with videos and images of all methods described, is available here: \url{https://tinyurl.com/rope-cloth-290t}.
\end{abstract}

\section{Simulator Design}
\begin{figure}
    \centering
    \includegraphics[width=0.65\textwidth]{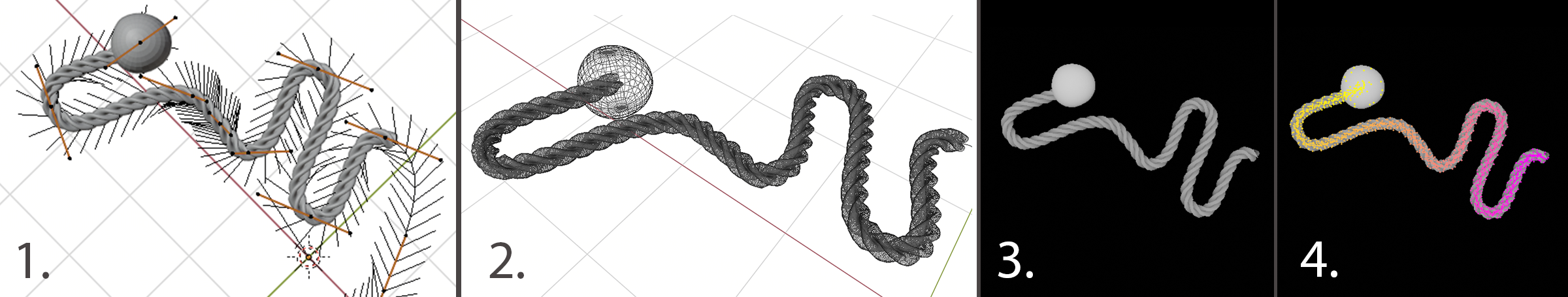}
    \caption{Blender simulation of rope 1. Bezier representation of rope with control points and handles (black), 2. Mesh view of rope, 3. Raw depth rendering of rope, 4. Rope with dense pixelwise ground truth annotations (colored according to indexing scheme) }
    \label{fig:my_label}
\end{figure}
We first worked on designing a simulation/rendering engine to generate datasets of the rope and cloth in varied configurations \cite{zhang2016health, zhang2020dex, zhang2021robots}. The main reason we chose to use a simulator for gathering data is that we require a large number of ground truth correspondence annotations to provide dense supervision, which is not readily available from real data without hand annotations \cite{devgon2020orienting, avigal20206, avigal2021avplug, sim2019personalization, elmquist2022art}. We also chose the simulator approach in order to see how various properties (texture, RGB, depth) affect correspondence quality, whereas to do the same in real, we would need to collect huge volumes of data, hand-annotate them, and risk overfitting to the specific datasets.

To render images of the rope, we use the open-source simulation and rendering software Blender 2.8, which has a convenient Python API \cite{pan2022tax, pan2023tax, eisner2022flowbot3d, zhang2023flowbot++, jin2024multi}. We modeled the rope as a mesh which is essentially a Bezier curve with a braided texture and volume. The control points of the Bezier curve can be randomized to produce arbitrary deformed configurations for rendering, or can also be geometrically structured into meaningful shapes like loops and knots to simulate shapes with self-occlusion \cite{yao2023apla, shen2024diffclip, lim2021planar, lim2022real2sim2real}. The rope mesh consists of ordered 3D vertices, and we can query their ground truth 2D pixel projects using the known camera-to-world transform in Blender.

To render images of the cloth in different states, we use the CS 184 mass-and-spring based cloth simulator available at \url{https://cs184.eecs.berkeley.edu/sp19/article/34/assignment-4-cloth-simulation} \cite{cs184} to generate 3D meshes of a textureless cloth in different "tiers" from flattened to heavily scrunched up. These meshes can be imported into Blender and rendered to give ground truth annotations similar to the rope setup.

\section{Problem Statement}
We seek to address the following problem:
Given an image of a deformable object $I_a$ with a source pixel $(u_a, v_a)$, find its correspondence $(u_b, v_b)$ in a different image $I_b$ of the same deformable object in a different configuration. 

To do this, we first map $I_a, I_b$ to an intermediary descriptor (feature) space given by the function $f$. $f$ is a pixel-wise mapping of a full-resolution RGB image to a $D$ -dimensional descriptor space: $(W \times H \times 3) \longmapsto (W \times H \times D)$.

We want to learn $f$ subject to the minimization $||f(I_b)[u_b, v_b] - f(I_a)[u_a, v_a]||_2$ for a given ground truth correspondence pair $(u_a, v_a), (u_b, v_b)$ in $I_a, I_b$. That is, corresponding pixels should be mapped to vectors that are close in descriptor space.

\begin{figure}[h!]
    \centering
    \includegraphics[width=6cm]{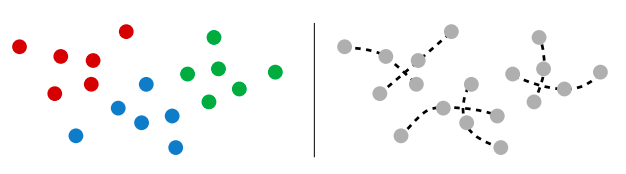}
    \caption{On the left is a visualization of the classification problem, where the objective is learning boundaries that separate clusters. On the right is the visualization of the task we are interested in — we are interested in learning correspondence within clusters. The function $f$ operates on members of each cluster, assigning semantic meaning to them. \cite{schmidt2016self}}
    \label{fig:my_label}
\end{figure}

Ideally, we would also like some metric of the confidence for a predicted correspondence, so we would like to compute: 
$p_b ((\hat{u_b}, \hat{v_b})| f, I_a, (u_a, u_v), I_b)$. This property is included in our proposed method for correspondence.

Each of the approaches outlined below proposes a different candidate for the $f$ function.

\section{Methods}

In the section, we summarize baseline correspondence methods that we implemented and used. Relevant code is either linked at the end of each subsection or sent over email. Our SIFT, SURF, and ORB implementations can be found at \url{https://tinyurl.com/siftsurforb290t}.

\subsection{SIFT: Scale Invariant Feature Transform}
For our first baseline, we implemented SIFT, Scale Invariant Feature Transform \cite{lowe2004sift}. SIFT takes an image and down-samples the image by to a quarter of the size for $n$ times. SIFT is a keypoint detector that leverages the Laplacian filter to find feature points, blurs and octave for scale-invariance, and the dominant orientation around a keypoint for rotation-invariance. For each image, SIFT uses a Gaussian blur at 5 or more different levels to generate a set of same-sized images at various blurs; these sets are known as octaves. SIFT then takes the Difference of Gaussian between consecutive layers in each octave to approximate the scale-invariant Laplacian of Gaussian. SIFT uses triplets of Difference of Gaussian images and calculates local maxima over subsets (these becomes your keypoints). We filter keypoints by intensity. For rotation invariance, SIFT uses magnitude and orientation to calculate the dominant orientation from any orientation that meets a threshold (0.8) for 4x4 pixel-blocks in a 16x16 pixel-area around the keypoint. Each keypoint is represented as a 128-dimension keypoint descriptor. For correspondence between 2 images that SIFT was used on, Lowe removes any points that closely matched to multiple points $(> 0.8)$. In \cite{lowe2004sift}, with this test, Lowe removed $90\%$ of false matches and only $5\%$ of correct matches. Our implementation can be found at .

\subsection{SURF: Speeded Up Robust Features}
For our second baseline, we implemented SURF, Speeded Up Robust Features \cite{bay2008surf}. Like SIFT, it is rotation and scale invariant. However, it more efficient at the loss of some accuracy. At its lowest scale, the authors approximate 2nd-order derivatives: $D_{xx}$, $D_{yy}$ and $D_{xy}$ with a box filter to calculate the determinant of the Hessian of $x, y$. Using larger box filters, they use the 2nd-order approximations from the lowest scale for other layers by using those approximations over larger box filters. The advantage of this is that we only need to calculate the second derivatives once, not at every scale like in SIFT. To achieve rotation invariance, the authors use the Haar Wavelet filter for a $6*s$, where $s$ is the scale factor) over intervals of $\pi/3$ and sum up the response. This is done for 16 subregions around the feature point. The keypoint descriptor is 64-dimension, for the 16 subregions and a 4D intensity (the Haar Wavelet response in the $dx$ and $dy$ directions.

\subsection{ORB: Oriented FAST and Rotated BRIEF}
We also implemented a baseline for correspondence using Oriented FAST and Rotated BRIEF (ORB) \cite{rublee2011orb}. ORB is a cheaper alternative to SIFT and SURF. ORB is not patented and free to use. Essentially, ORB is a fusion of FAST and BRIEF: it uses FAST to find keypoints of the given pair of pictures, and then it applies Harris corner measure to find top $N$ points among them. ORB restores orientation of FAST by computing the intensity weighted centroid of the patch with located corner at center. The direction of the vector from this corner point restores the orientation. Then ORB uses BRIEF descriptor to match correspondence, and the novelty is that it uses the orientation computed in the previous step to improve the performance of BRIEF descriptor on rotation.
\subsection{TimeCycle}
For the first deep learning approach to correspondence matching, we tried an approach called Time Cycle. In \cite{wang2019learning}, the authors noticed that given a video, to match the correspondences in the video, one can track a pixel back to older frames, and then track the pixel from the older frames back to the current frames. Therefore, one can try to minimize the discrepancy of these pixel locations as a loss function. This completed a cycle of correspondence tracking, and one must enforce the cycle consistency in order to improve correspondence learning. To make the learning more robust, one can optionally skip frames during training time because tracking consecutive frames is sometimes not challenging enough. Consider inputs as a sequence of video frames $I_{t-k:t}$ and a patch $p_t$ taken from $I_t$. These pixel inputs are mapped to a feature space by an encoder $\phi$, such that $x^I_{t-k:t} = \phi(I_{t-k:t})$
and $x^p_t = \phi(p_t).$ Therefore, the loss functions are defined as follows:

First we define a cycle consistency loss. Given a tracker function $\mathcal{T}$ that tracks the patch features, we define forward tracking $i$ time steps as $\mathcal{T}^{(i)}$, and we define backward tracking $i$ time steps as $\mathcal{T}^{(-i)}$. We are also given patch features at time step $t$ as $x_t^p$ and image features at time step $s$ as $x_s^I$, we use $\mathcal{T}$ to find patch features at time step $s$ that is most similar to $x_t^p$. The cycle consistency loss is thus defined:
$$\mathcal{L}^i_{long} = l_\theta(x_t^p, \mathcal{T}^{(i)}(x^I_{t-i+1}, \mathcal{T}^{(-i)}(x_{t-1}^I,x_t^p)))$$
Then we define a skip cycle loss, where we can optionally skip $i$ steps:
$$\mathcal{L}^i_{skip} = l_\theta(x_t^p, \mathcal{T}(x_t^I, \mathcal{T}(x_{t-i}^I,x_t^p)) $$
We also define a similarity loss. The idea is to explicitly check the similarity between the query patch feature $x_t^p$ and the tracked patch $\mathcal{T}(x_{t-i}^I,x_t^p)$ in feature space. \cite{wang2019learning} defines the similarity loss as the negative Frobenius inner product of the two feature tensors:
$$\mathcal{L}^i_{sim} = -\langle x_t^p, \mathcal{T}(x_{t-i}^I,x_t^p)\rangle$$
The $l_\theta$ operator used above is defined to gauge the alignment between patches, and it is the sum of L2 norm of the difference between the bilinearly sampled patches. Thus, the overall loss function is the weighted sum of the three types of losses, summed over $k$ cycles. The original TimeCycle network was trained on DAVIS-2017 dataset. The videos in that dataset are mostly full of features, so it is relatively easier to track the features in such videos. However, we would like to try to learn correspondence matching using TimeCycle on deformable 1D objects such as ropes. Ropes are more challenging because they are highly deformable and, in most cases, featureless. Therefore, we generated our own rope dataset by simulating random an action on the rope in each frame, and rendered the resulting image. All the images were rendered using Blender 2.80, and we used the rendered images to retrain TimeCycle. The dataset is consisted of 250 episodes where each episode contains 3000 images, and it took about a day to render. Then we retrained the network using the new rope dataset.

\subsection{Dense Object Nets - Pixelwise Contrastive Method}

This method of learned correspondence \cite{dense-obj-nets} \cite{schmidt2016self} learns a pixel-wise mapping of a full-resolution RGB image to a $D$ -dimensional descriptor space: $(W \times H \times 3) \longmapsto (W \times H \times D)$. During training, a pair of images is sampled in a Siamese fashion, and sets of both matching pixels and non-matching pixels are sampled within the image pair. In this formulation, the objective is divided into two parts, similar to a triplet loss. 

First, it seeks to minimize the distance between matching correspondence-pairs: 
$$||f(I_b)[u_b, v_b] - f(I_a)[u_a, v_a]||_2$$ for a given ground truth correspondence pair $(u_a, v_a), (u_b, v_b)$ in $I_a, I_b$. Second, it enforces that non-matches (arbitrarily sampled pixels that have no correspondence) should be pushed apart in descriptor space by a fixed margin $M$.

Formally, the loss is defined as follows\cite{schmidt2016self}:

$L(I_a, I_b, u_a, u_b)$ =
\[ \begin{cases} 
      ||f(I_b)[u_b, v_b] - f(I_a)[u_a, v_a]||_2 & (u_a, v_a), (u_b, v_b)  \textnormal{ is a correspondence pair} \\
      \textnormal{max}(0, M - ||f(I_b)[u_b, v_b] - f(I_a)[u_a, v_a]||_2)^2 & (u_a, v_a), (u_b, v_b) \textnormal{ is not a correspondence pair} 
   \end{cases}
\] 

It is important to note that the original authors of the Dense Object Nets paper only tried this method on real images of rigid objects. In their case, they used knowledge of the camera intrinsics and camera-to-world transformation as a supervisory signal during training for labelling whether two pixels corresponded (based on their deprojection to world space). We are able to generalize the method to synthetic images of deformable objects using our simulator's ground truth as a supervisory signal. 

\subsection{Proposed Method - Enforcing Spatial Continuity in Correspondence Estimation Through a Reformulated Loss Function}
While Pixelwise Contrastive Loss proves to be an effective loss function for our datasets, when provided with noise injected images, it is prone to large discontinuous "jumps" in correspondence estimation. We hypothesize that these discontinuities are a result of a lack of enforced spatial continuity during training time  which is a natural pitfall of pixelwise contrastive loss. Another issue with the Vanilla Dense Object Nets formulation is that the hyperparameter $M$ is a bit arbitrarily chosen, and enforces the same penalty (that non-matches should be pushed apart by a margin) regardless of the severity of the non-match. Instead, we want to enforce a spatial continuity which yields a relative interpretation of how good or poor a given match is considering its surrounding region. We propose two new formulations of the loss function: distributional loss and L-Lipschitz loss which implicitly and explicitly, respectively, enforce spatial continuity across nearby correspondence estimations. 

\subsection{Distributional Loss}
Code: \url{https://tinyurl.com/distrib-loss}

The idea here is that instead of using a contrastive approach, we merely want to learn $f$ such that $p_a ((\hat{u_a}, \hat{v_a})| f, I_b, (u_b, v_b), I_a)$ matches a ground truth distribution, $q_a$, which should enforce spatial continuity around the ground truth correspondence. To implicitly enforce spatial continuity, we want the ground truth distribution to look like a Gaussian centered at the actual correspondence, with probability mass spreading outwards according to the chosen $\sigma$. $p_a$ can be thought of as the distribution fit so far over the whole of $I_a$ predicting the per-pixel probability of correspondence for $(u_b, v_b)$. We can compute it by taking the softmax over descriptor norm differences, comparing each pixel in $I_a$'s descriptor $f(I_a)[\hat{u_a}, \hat{v_a}]$ to $f(I_b)[u_b, v_b]$.

\begin{figure}[h!]
    \centering
    \includegraphics[width=10cm]{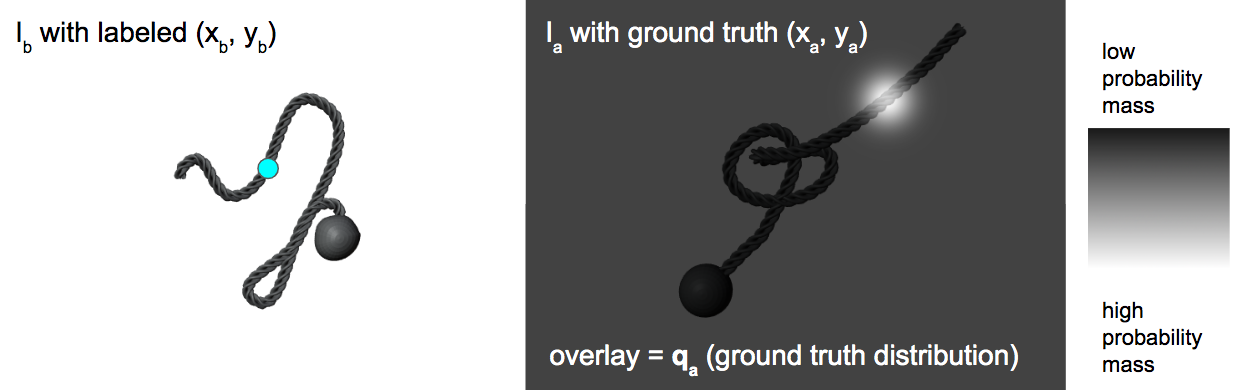}
    \caption{Ground truth $q_a$ distribution.}
    \label{fig:my_label}
\end{figure}

This formulation provides flexibility in choosing our $q_a$ distribution, and we can also resolve the symmetry issue and do away with the ball by splitting probability mass symmetric about the length of the rope (bimodal distribution). This would allow us to predict the symmetric correspondences for a pixel \emph{separately}.

\begin{figure}[h!]
    \centering
    \includegraphics[width=3cm]{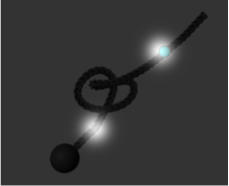}
    \caption{Ground truth bimodal $q_a$ distribution.}
    \label{fig:my_label}
\end{figure}

\subsection{L-Lipschitz Regularization}
Code: \url{https://tinyurl.com/lipschitz-loss}
We propose to explicitly enforce spatial continuity during training time by adding an additional regularization term to the distributional loss based on the following inequality:
$$\underbrace{\left | \left | 
\begin{bmatrix}
\hat x_{a, 1} \\
\hat y_{a,1}
\end{bmatrix} 
- \begin{bmatrix}
\hat x_{a, 2} \\
\hat y_{a,2}
\end{bmatrix}
\right | \right|_2
}_{L2 \text{ pixel distance in \textbf{source} image}}
\leq L\underbrace{\left | \left | 
\begin{bmatrix}
\hat x_{b, 1} \\
\hat y_{b,1}
\end{bmatrix} 
- \begin{bmatrix}
\hat x_{b, 2} \\
\hat y_{b,2}
\end{bmatrix}
\right | \right|_2}_{L2 \text{ pixel distance in \textbf{target} image}}
$$

In the inequality above, the term on the left represents the L2 distance in pixel space between to arbitrary pixels in the source image.  When both these pixels are mapped into descriptor space, their respective correspondences in the target image are determined by finding the pixels in the target image whose values in descriptor space are closest to the descriptor values of the source pixels.  We will refer to the corresponding pixels in the target image as the "best match" pixels.  The norm on the right of the inequality represents the L2 distance in pixel space between the best match target pixel for the first source pixel and the best match target pixel for the second source pixel.  L here is simply a scaling factor.

Intuitively, the inequality states that pixels that are d units apart in the source image should be no more than L*d units apart in the target image.  In order to formalize the inequality into a regularization penalty, we simply subtract the right hand side from both sides of the inequality and apply ReLu for non-negativity:
$$\underbrace{\mu \cdot \max\left(0, \underbrace{\left | \left | 
\begin{bmatrix}
\hat x_{a, 1} \\
\hat y_{a,1}
\end{bmatrix} 
- \begin{bmatrix}
\hat x_{a, 2} \\
\hat y_{a,2}
\end{bmatrix}
\right | \right|_2
}_{L2 \text{ pixel distance in \textbf{source} image}}
- \;\;\; L\underbrace{\left | \left | 
\begin{bmatrix}
\hat x_{b, 1} \\
\hat y_{b,1}
\end{bmatrix} 
- \begin{bmatrix}
\hat x_{b, 2} \\
\hat y_{b,2}
\end{bmatrix}
\right | \right|_2}_{L2 \text{ pixel distance in \textbf{target} image}} \right )}_{\text{Lipschitz Penalty}}
$$

$\mu$ is a constant that scales the penalty and usually starts small and increases gradually during training. Additionally, in theory the penalty would be enforced on all pairwise pixels in the source image, however, due to computational and memory constraints, we sample a patch around a set of pixels for which we have ground truth information during training.  This vastly reduces the required memory and training time.

\subsection{Enforcing Time-Consistency}
Interestingly enough, the same L-Lipschitz regularization penalty from above, only slighty modified, can be used to explicitly enforce temporal continuity, a factor that is also lacking in the Vanilla Dense Object Nets formulation. Instead of applying a constraint to pairwise pixels in the source image and their relative locations in the target image, we can apply a constraint directly between a pixel in the source image and its corresponding location in the target image as follows:
$$\underbrace{\mu \cdot \max\left(0, \underbrace{\left | \left | 
\begin{bmatrix}
\hat x_{a, 1} \\
\hat y_{a,1}
\end{bmatrix} 
- \begin{bmatrix}
\hat x_{a, 2} \\
\hat y_{a,2}
\end{bmatrix}
\right | \right|_2
}_{L2 \text{ pixel distance across single correspondence in \textbf{source} image}}
- \;\;\; L\underbrace{\left | \left | 
\begin{bmatrix}
\hat x_{b, 1} \\
\hat y_{b,1}
\end{bmatrix} 
- \begin{bmatrix}
\hat x_{b, 2} \\
\hat y_{b,2}
\end{bmatrix}
\right | \right|_2}_{L2 \text{ time constraint}} \right )}_{\text{Lipschitz Temporal Penalty}}
$$

If we make the assumption that we have access to some time-limited value such as the maximum distance and object can travel between consecutive frames, then we can use this information to enforce the penalty.  We have left this as future work.

\section{Overall Quantitative Results}
\begin{table}[!hbtp]
  \caption{RMSE of Correspondences}
  \label{sample-table}
  \centering
  \begin{tabular}{|l|l|l|l|l|l|}
    \toprule
    % \multicolumn{2}{c}{Correspondence Error (RMSE)} \\
    \cmidrule(r){1-2}
    Dataset & SIFT & SURF & ORB & DON (Contrastive Loss) & DON (Distribution Loss)\\
    % \midrule
    Rope & 104 & 95 & 93 & 28.5 & 31.27  \\
    Cloth Tier 1 & 162 & 161 & 178 & 25.5 & 32.4     \\
    Cloth Tier 3 & 181 & 190 & 200 & 33.4 & 41.5 \\
    \bottomrule
  \end{tabular}
\end{table}

\section{Detailed Results/Appendix}
We used OpenCV to calculate keypoints and find the corresponding keypoints for consecutive pairs of images with SIFT, SURF, and ORB. The images below show the top 10 keypoint correspondences according to the quality of feature match between two keypoints, not according to the ground truth correspondences. We plotted all the keypoints that a given baseline outputted for rope, tier 1 cloth, and tier 3 cloth. The histograms are normalized over the range of RMSE values. We expect normalized distribution to have a mean over 0.5, otherwise it is skewed. We have also noted the unnormalized range (in pixels) in the figure description for each baseline/object pairing. We evaluate the histograms by looking at the normalized pair distances (randomly paired) (positively skewed indicates consistency, over 0.5 indicates a normal distribution, negatively skewed indicates inconsistency) and then at the normalized RMSE values (the larger they are, the less accurate the pixel correspondences). The code can be found at \url{https://tinyurl.com/rmse290t}.

We also plotted histogram for RMSE frequency across correspondence pairs, which are the two histograms below the rope images. 

 We were also curious at how many correspondences were within a range of l2 distance from the ground truth. The l2 distance histogram shows correspondence errors vs the frequency at which they occur. In the rope datasets, the majority of error is within 50 pixels and the histogram is skewed right, while for the cloth dataset, the histogram looks more normal and the majority of correspondences have errors between 100 and 200.
\subsection{SIFT}

\begin{figure}[!htbp]%
    \centering
    \subfloat[]{{\includegraphics[width=0.9\textwidth]{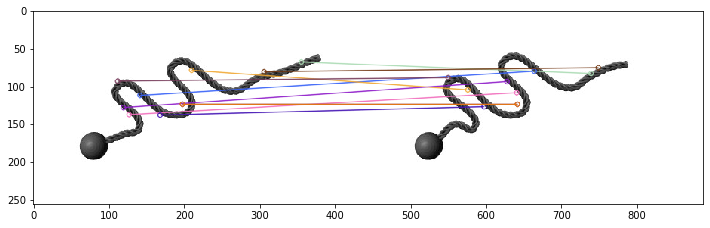} }}%
    \;
    \subfloat[]{{\includegraphics[width=0.52\textwidth]{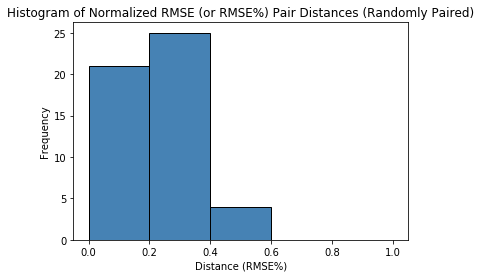} }}%
    \;
    \subfloat[]{{\includegraphics[width=0.4\textwidth]{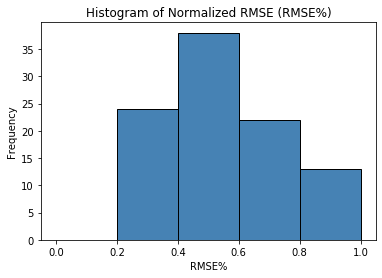} }}%
    \;
    \caption{SIFT Correspondence on rope dataset. From (b), the histogram shows the error is consistent between RMSE pair distances. However, from (c), we can see that said errors are often large and have an unnormalized range from 39 to 223 pixels. A box-and-whisker plot (not shown) of unnormalized RMSE shows 223 is an outlier. SIFT seems unable to distinguish the local area of a keypoint from other areas of the rope.}%
    \label{fig:example}%
\end{figure}

\begin{figure}[!htbp]%
    \centering
    \includegraphics[width=0.7\textwidth]{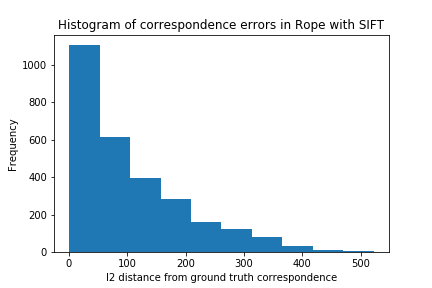}
    \caption{Histogram of L2 distance of SIFT Correspondences from ground truth correspondences on Rope dataset }%
    \label{fig:example}%
\end{figure}

\begin{figure}[!htbp]%
    \centering
    \includegraphics[width=0.48\textwidth]{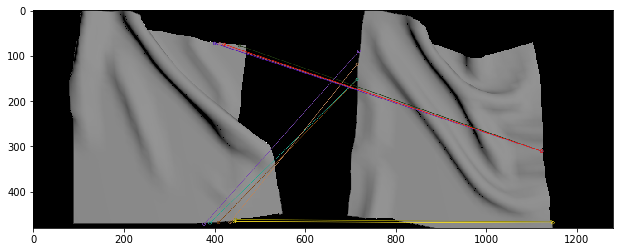}
    \includegraphics[width=0.48\textwidth]{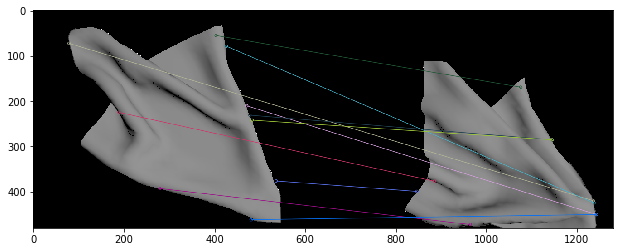}
    \caption{SIFT Correspondence on tier 1 and tier 3 cloth dataset. On texture-less objects, SIFT struggles as expected to find features. In contrast to our previous robotic work (tier 1 smooth is "easier" for a robot to smooth that tier 3), the wrinkles and hidden corners in tier 3 allow it to perform better correspondence.}%
    \label{fig:example}%
\end{figure}

\begin{figure}[!htbp]%
    \centering
    \includegraphics[width=0.48\textwidth]{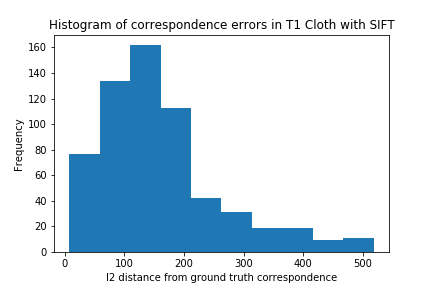}
    \includegraphics[width=0.48\textwidth]{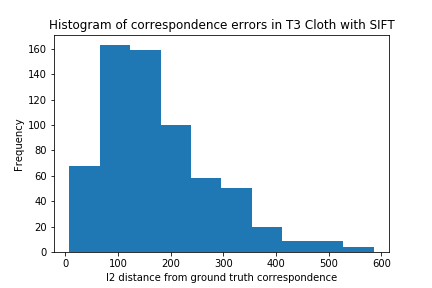}
    \caption{Histogram of L2 distance of SIFT Correspondences from ground truth correspondences on Tier 1 and Tier 3 Cloth Dataset}%
    \label{fig:example}%
\end{figure}

\subsection{SURF}
\begin{figure}[!htbp]%
    \centering
    \subfloat[]{{\includegraphics[width=0.9\textwidth]{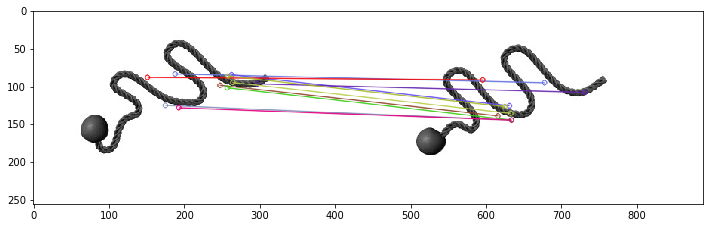} }}%
    \;
    \subfloat[]{{\includegraphics[width=0.54\textwidth]{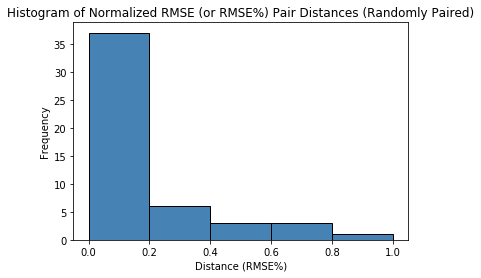} }}%
    \;
  \subfloat[]{{\includegraphics[width=0.42\textwidth]{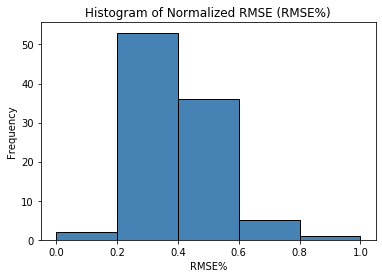} }}%
    \;
    \caption{SURF Correspondence on rope dataset. The RMSE\% pair-distance histogram (b) shows that the data was consistent. The RMSE\% histogram (c) indicates that most points were accurate. However, there are a lot of outliers that have really high errors. The unnormalized range is [37, 272] pixels, which is larger than SIFT's on rope, despite performing better on the majority of correspondences.}%
    \label{fig:example}%
\end{figure}

\begin{figure}[!htbp]%
    \centering
    \includegraphics[width=0.7\textwidth]{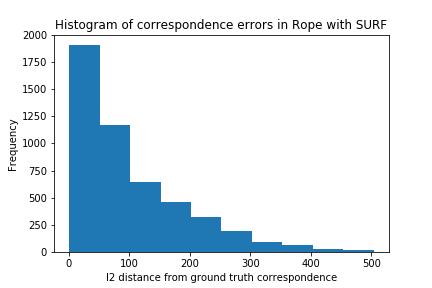}
    \caption{Histogram of L2 distance of SURF Correspondences from ground truth correspondences on Rope dataset }%
    \label{fig:example}%
\end{figure}

\begin{figure}[!hbtp]%
    \centering
    \includegraphics[width=0.48\textwidth]{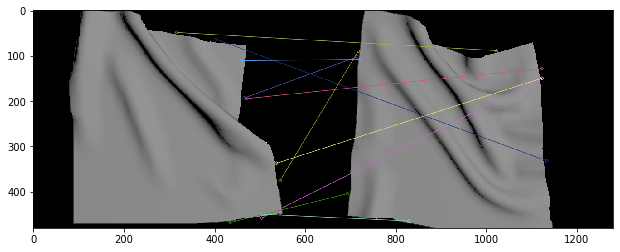}
    \includegraphics[width=0.48\textwidth]{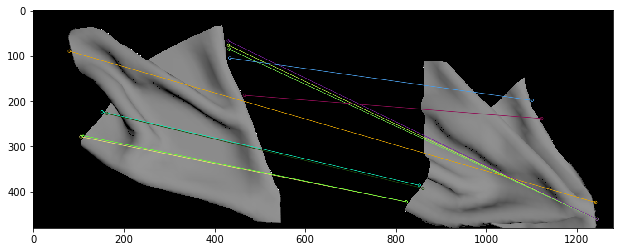}
    \caption{SURF Correspondence on tier 1 and tier 3 cloth dataset.}%
    \label{fig:example}%
\end{figure}

\begin{figure}[!htbp]%
    \centering
    \includegraphics[width=0.48\textwidth]{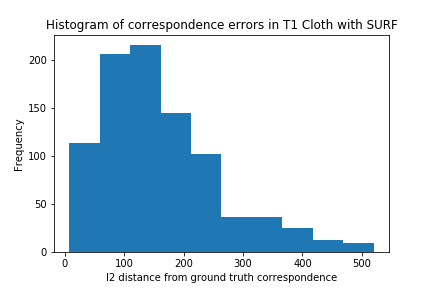}
    \includegraphics[width=0.48\textwidth]{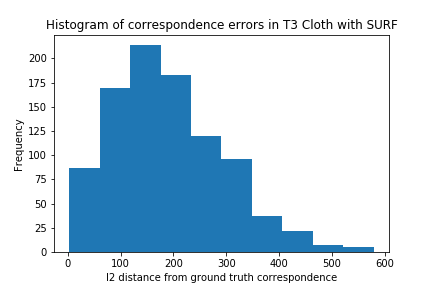}
    \caption{Histogram of L2 distance of SURF Correspondences from ground truth correspondences on Tier 1 and Tier 3 Cloth dataset}%
    \label{fig:example}%
\end{figure}

\subsection{ORB}
\begin{figure}[!hbtp]%
    \centering
    \subfloat[]{{\includegraphics[width=0.9\textwidth]{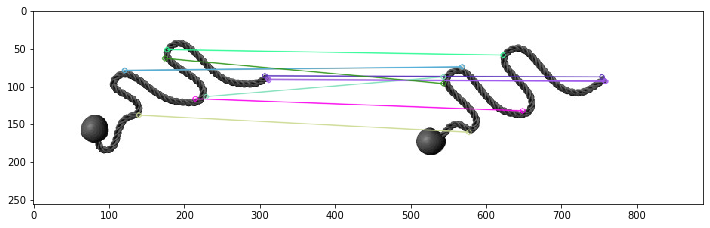} }}%
    \;
    \subfloat[]{{\includegraphics[width=0.54\textwidth]{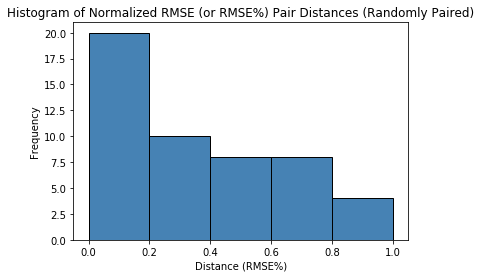} }}%
    \;
  \subfloat[]{{\includegraphics[width=0.42\textwidth]{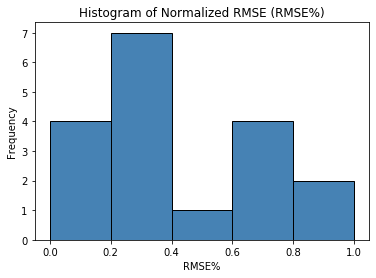} }}%
    \;
    \caption{ORB Correspondence on rope dataset. The RMSE\% Pair-distances histogram (b) indicates less consistent values than in SIFT and SURF. The RMSE\% histogram (c) is bimodal which is much less accurate than the other methods despite having no outliers (because of the larger interquartile range) and qualitatively looking like it performed the best.}%
    \label{fig:example}%
\end{figure}

\begin{figure}[!htbp]%
    \centering
    \includegraphics[width=0.7\textwidth]{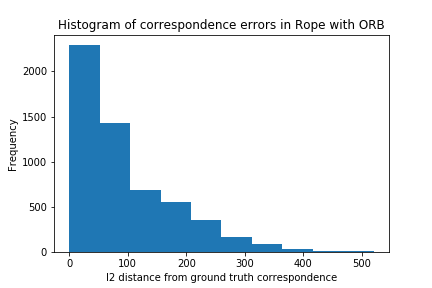}
    \caption{Histogram of L2 distance of ORB Correspondences from ground truth correspondences on Rope dataset}%
    \label{fig:example}%
\end{figure}

\begin{figure}[h!]%
    \centering
    \includegraphics[width=0.48\textwidth]{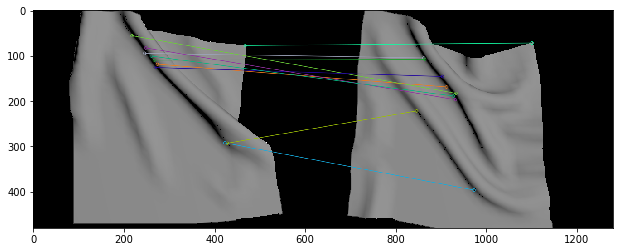}
    \includegraphics[width=0.48\textwidth]{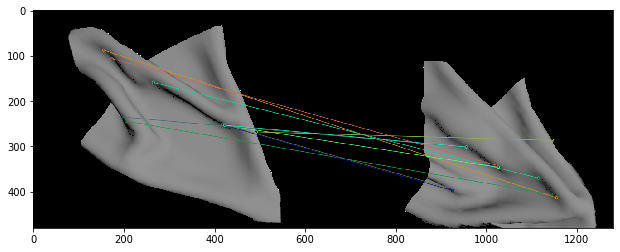}
    \caption{ORB Correspondence on tier 1 and tier 3 cloth dataset. }%
    \label{fig:example}%
\end{figure}

\begin{figure}[!htbp]%
    \centering
    \includegraphics[width=0.48\textwidth]{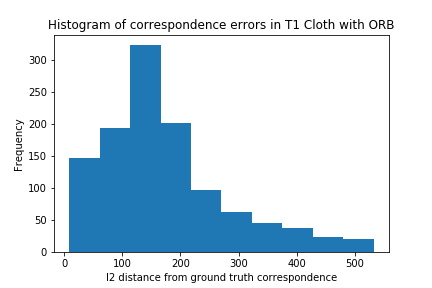}
    \includegraphics[width=0.48\textwidth]{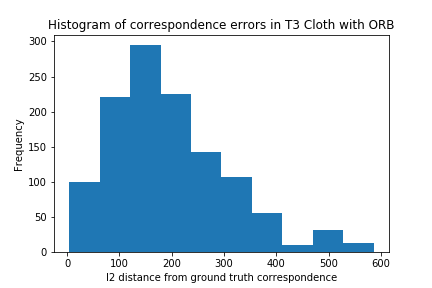}
    \caption{Histogram of L2 distance of ORB Correspondences from ground truth correspondences on Tier 1 and Tier 3 Cloth dataset}%
    \label{fig:example}%
\end{figure}

\subsection{TimeCycle}
\begin{figure}[!hbtp]
    \centering
    \begin{tabular}{c@{\hspace{10pt}}c@{\hspace{10pt}}c@{\hspace{10pt}}}
    \subfloat[Ground truth]{\includegraphics[width=45mm, clip]{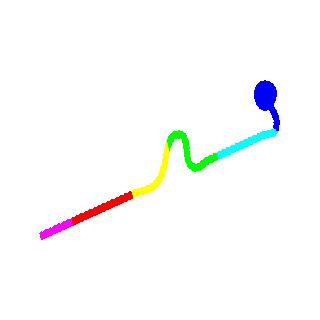}}&
    \subfloat[Intermediary correspondence matching]{\includegraphics[width=40mm, clip]{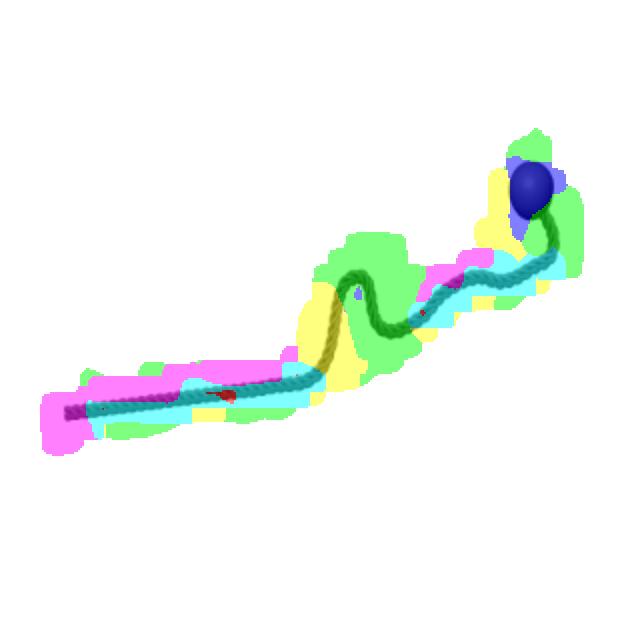}}&
    \subfloat[Final correspondence matching]{\includegraphics[width=40mm, clip]{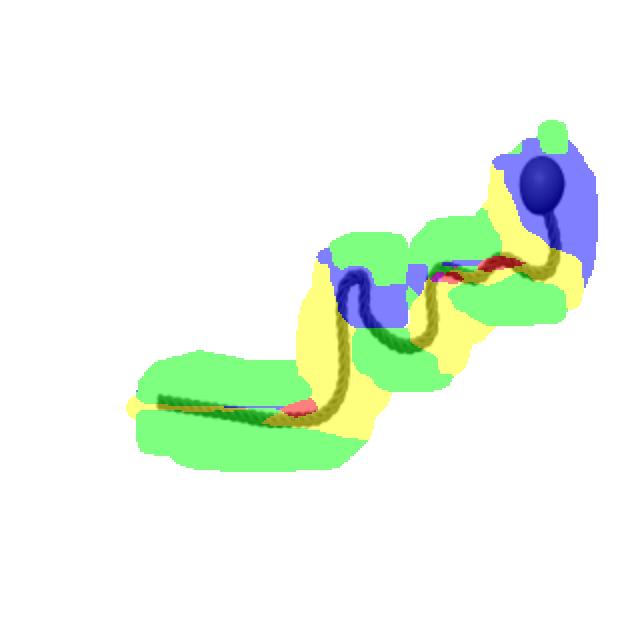}}
    \end{tabular}
    \caption{TimeCycle: We tested our retrained network using a small test dataset provided with annotations. We would like to see how well the network tracks the provided features across different frames. Here we show the result of Time Cycle tracking rope's correspondence. \textbf{Left:} Ground truth label (provided). \textbf{Middle:} Correspondence after 22 frames. \textbf{Right:} Correspondence after 87 frames.}
    \label{fig:timecycle}
\end{figure}

\subsection{Vanilla Dense Object Nets}
\begin{figure}[!hbtp]%
    \centering
    \subfloat[]{{\includegraphics[width=4cm]{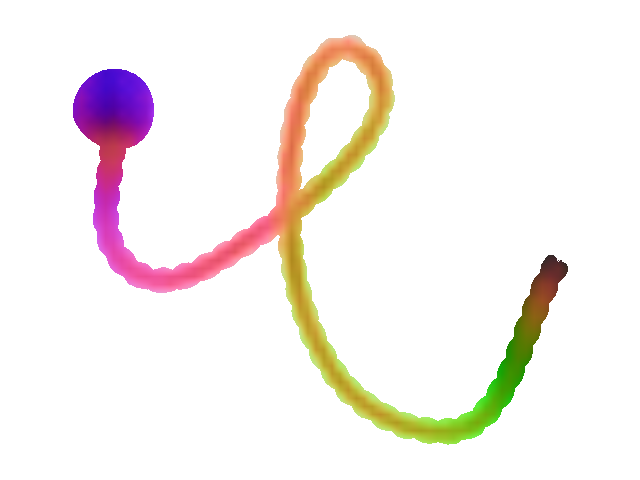} }}%
    \qquad
    \subfloat[]{{\includegraphics[width=4cm]{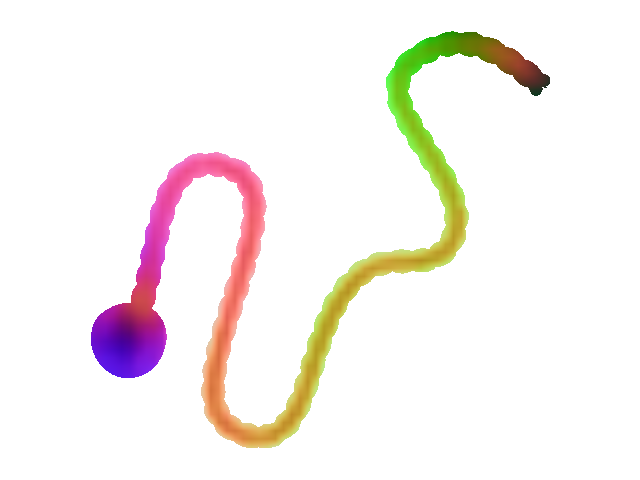} }}%
    \qquad
   \subfloat[]{{\includegraphics[width=4cm]{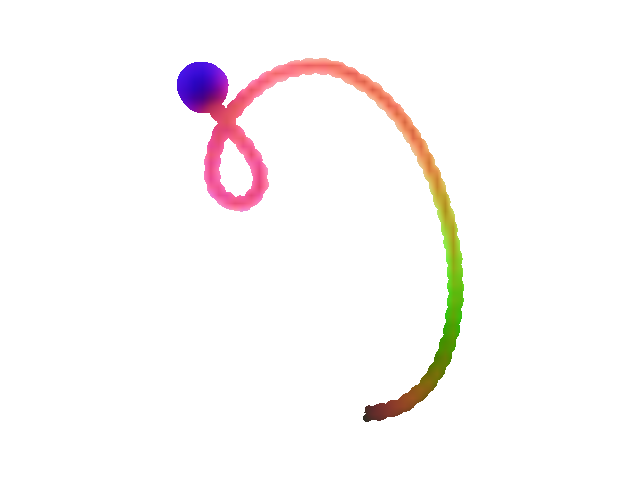} }}%
    \qquad
    \caption{Vanilla Dense Object Nets: We show a masked visualization of the results of mapping images of rope in different configurations to 3 dimensional descriptor space, visualized as R,G,B tuples for each descriptor dimension. We see that the descriptors for the most part yield semantically consistent correspondence, although they still struggle with self-occlusion and loops as in example (a), where the orange and yellow regions are improperly ordered.}%
    \label{fig:example}%
\end{figure}
\begin{figure}[!hbtp]%
    \centering
    \subfloat[]{\includegraphics[width=6cm]{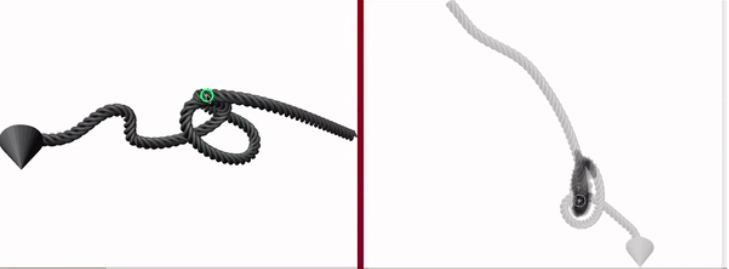} }%
    \qquad
    \subfloat[]{\includegraphics[width=6cm]{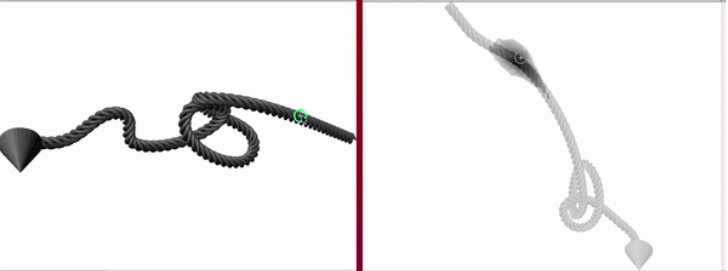} }%
    \qquad
    \caption{Vanilla Dense Object Nets. Correspondence heatmaps for a given source pixel (green circle) and a different image. For every pixel in the target image (right in each pair), we take the L2 norm of that pixel's descriptor with the source pixel's descriptor and scale this linearly $\in$ [0, 255]. The best match for the source pixel is the argmin of this L2 norm descriptor difference. }%
    \label{fig:example}%
\end{figure}

\begin{figure}[!htbp]
  \centering
  \begin{tabular}{@{}c@{}}
    \includegraphics[width=.6\linewidth]{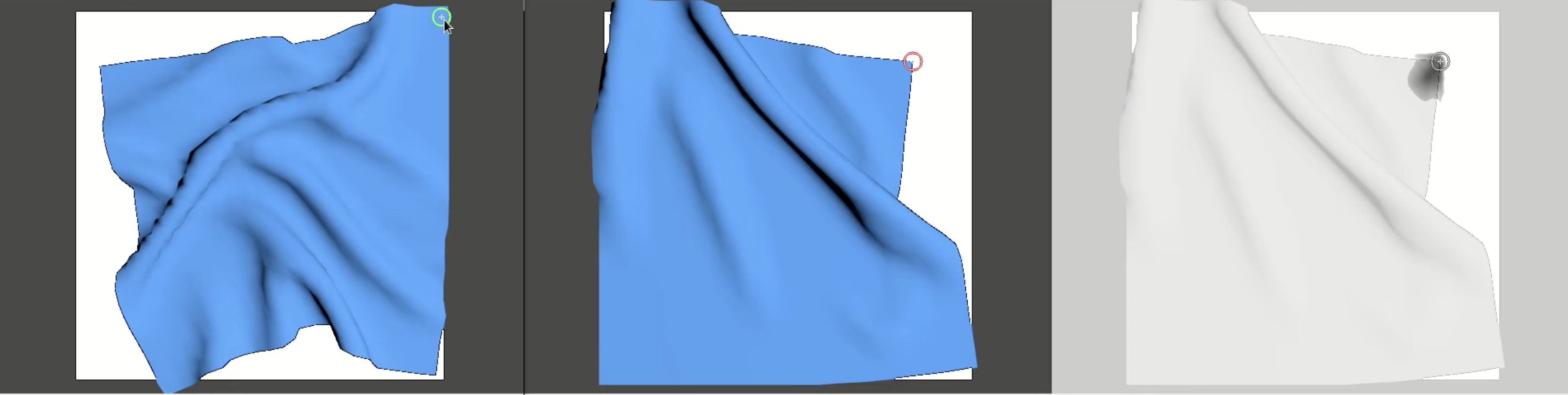} \\[\abovecaptionskip]
    \small (a) Correspondence heatmap for a simulated cloth color image.
  \end{tabular}
  \vspace{\floatsep}
  \begin{tabular}{@{}c@{}}
    \includegraphics[width=.6\linewidth]{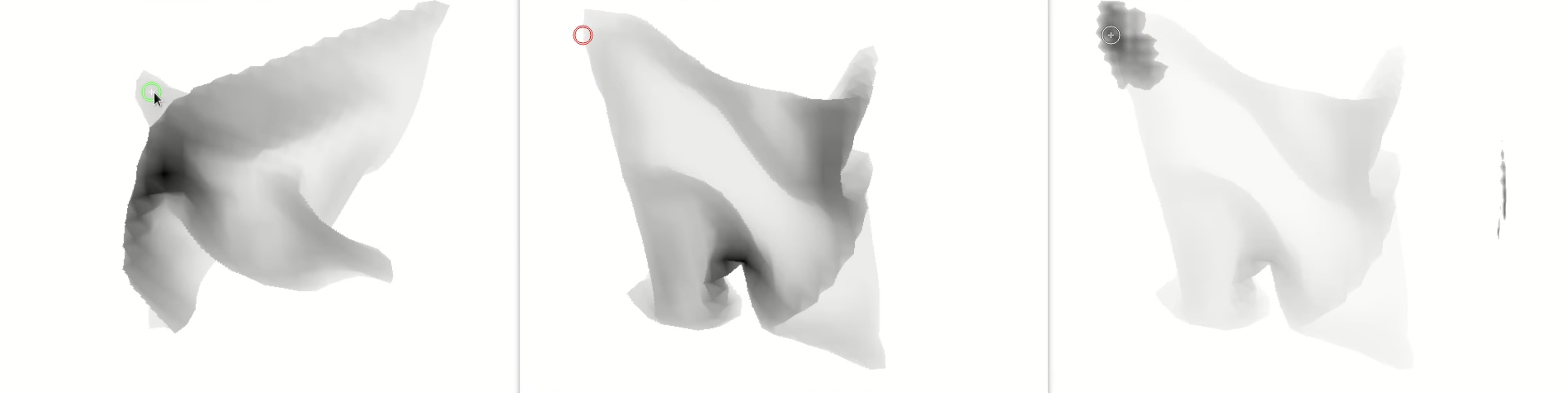} \\[\abovecaptionskip]
    \small (b) Correspondence heatmap for a simulated cloth depth image.
  \end{tabular}
  \begin{tabular}{@{}c@{}}
    \includegraphics[width=.6\linewidth]{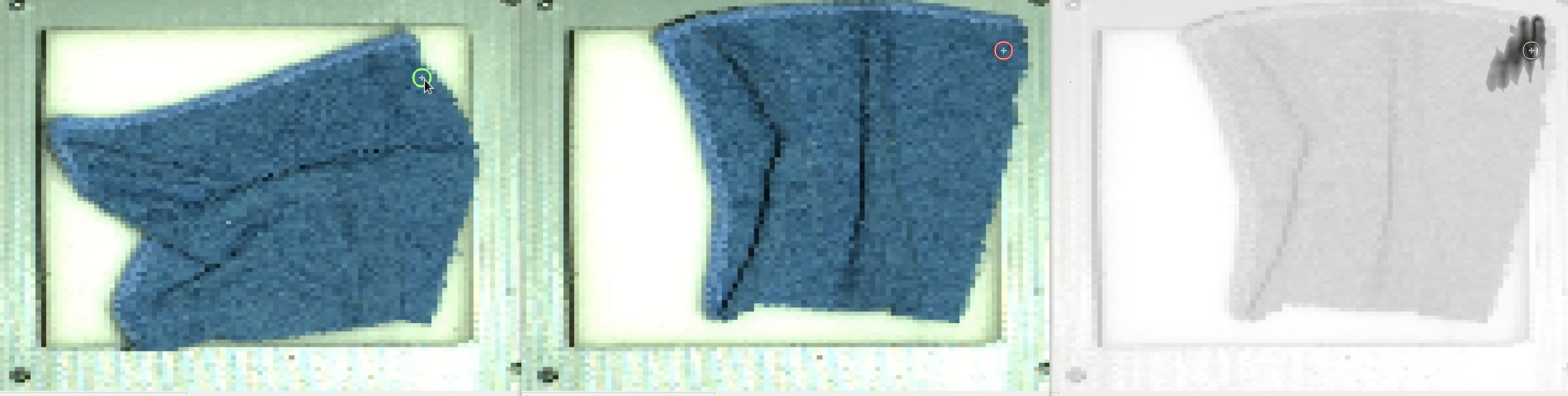} \\[\abovecaptionskip]
    \small (b) Correspondence heatmap for a real cloth color image.
  \end{tabular}
  \caption{Vanilla Dense Object Nets: Correspondence heatmaps for images of simulated and real cloth using a dense object net model trained with pixelwise contrastive loss.}\label{fig:myfig}
\end{figure}

\bibliographystyle{IEEEtran}
\bibliography{ref.bib}

% Generated by IEEEtran.bst, version: 1.13 (2008/09/30)
\begin{thebibliography}{10}
\providecommand{\url}[1]{#1}
\csname url@samestyle\endcsname
\providecommand{\newblock}{\relax}
\providecommand{\bibinfo}[2]{#2}
\providecommand{\BIBentrySTDinterwordspacing}{\spaceskip=0pt\relax}
\providecommand{\BIBentryALTinterwordstretchfactor}{4}
\providecommand{\BIBentryALTinterwordspacing}{\spaceskip=\fontdimen2\font plus
\BIBentryALTinterwordstretchfactor\fontdimen3\font minus \fontdimen4\font\relax}
\providecommand{\BIBforeignlanguage}[2]{{%
\expandafter\ifx\csname l@#1\endcsname\relax
\typeout{** WARNING: IEEEtran.bst: No hyphenation pattern has been}%
\typeout{** loaded for the language `#1'. Using the pattern for}%
\typeout{** the default language instead.}%
\else
\language=\csname l@#1\endcsname
\fi
#2}}
\providecommand{\BIBdecl}{\relax}
\BIBdecl

\bibitem{zhang2016health}
H.~Zhang, ``Health diagnosis based on analysis of data captured by wearable technology devices,'' \emph{International Journal of Advanced Science and Technology}, vol.~95, pp. 89--96, 2016.

\bibitem{zhang2020dex}
H.~Zhang, J.~Ichnowski, Y.~Avigal, J.~Gonzales, I.~Stoica, and K.~Goldberg, ``Dex-net ar: Distributed deep grasp planning using a commodity cellphone and augmented reality app,'' in \emph{2020 IEEE International Conference on Robotics and Automation (ICRA)}.\hskip 1em plus 0.5em minus 0.4em\relax IEEE, 2020, pp. 552--558.

\bibitem{zhang2021robots}
H.~Zhang, J.~Ichnowski, D.~Seita, J.~Wang, H.~Huang, and K.~Goldberg, ``Robots of the lost arc: Self-supervised learning to dynamically manipulate fixed-endpoint cables,'' in \emph{2021 IEEE International Conference on Robotics and Automation (ICRA)}.\hskip 1em plus 0.5em minus 0.4em\relax IEEE, 2021, pp. 4560--4567.

\bibitem{devgon2020orienting}
S.~Devgon, J.~Ichnowski, A.~Balakrishna, H.~Zhang, and K.~Goldberg, ``Orienting novel 3d objects using self-supervised learning of rotation transforms,'' in \emph{2020 IEEE 16th International Conference on Automation Science and Engineering (CASE)}.\hskip 1em plus 0.5em minus 0.4em\relax IEEE, 2020, pp. 1453--1460.

\bibitem{avigal20206}
Y.~Avigal, S.~Paradis, and H.~Zhang, ``6-dof grasp planning using fast 3d reconstruction and grasp quality cnn,'' \emph{arXiv preprint arXiv:2009.08618}, 2020.

\bibitem{avigal2021avplug}
Y.~Avigal, V.~Satish, Z.~Tam, H.~Huang, H.~Zhang, M.~Danielczuk, J.~Ichnowski, and K.~Goldberg, ``Avplug: Approach vector planning for unicontact grasping amid clutter,'' in \emph{2021 IEEE 17th international conference on automation science and engineering (CASE)}.\hskip 1em plus 0.5em minus 0.4em\relax IEEE, 2021, pp. 1140--1147.

\bibitem{sim2019personalization}
K.~C. Sim, F.~Beaufays, A.~Benard, D.~Guliani, A.~Kabel, N.~Khare, T.~Lucassen, P.~Zadrazil, H.~Zhang, L.~Johnson \emph{et~al.}, ``Personalization of end-to-end speech recognition on mobile devices for named entities,'' in \emph{2019 IEEE Automatic Speech Recognition and Understanding Workshop (ASRU)}.\hskip 1em plus 0.5em minus 0.4em\relax IEEE, 2019, pp. 23--30.

\bibitem{elmquist2022art}
A.~Elmquist, A.~Young, T.~Hansen, S.~Ashokkumar, S.~Caldararu, A.~Dashora, I.~Mahajan, H.~Zhang, L.~Fang, H.~Shen \emph{et~al.}, ``Art/atk: A research platform for assessing and mitigating the sim-to-real gap in robotics and autonomous vehicle engineering,'' \emph{arXiv preprint arXiv:2211.04886}, 2022.

\bibitem{pan2022tax}
C.~Pan, B.~Okorn, H.~Zhang, B.~Eisner, and D.~Held, ``Tax-pose: Task-specific cross-pose estimation for robot manipulation,'' \emph{arXiv preprint arXiv:2211.09325}, 2022.

\bibitem{pan2023tax}
------, ``Tax-pose: Task-specific cross-pose estimation for robot manipulation,'' in \emph{Conference on Robot Learning}.\hskip 1em plus 0.5em minus 0.4em\relax PMLR, 2023, pp. 1783--1792.

\bibitem{eisner2022flowbot3d}
B.~Eisner, H.~Zhang, and D.~Held, ``Flowbot3d: Learning 3d articulation flow to manipulate articulated objects,'' \emph{arXiv preprint arXiv:2205.04382}, 2022.

\bibitem{zhang2023flowbot++}
H.~Zhang, B.~Eisner, and D.~Held, ``Flowbot++: Learning generalized articulated objects manipulation via articulation projection,'' \emph{arXiv preprint arXiv:2306.12893}, 2023.

\bibitem{teng2024gmkf}
S.~Teng, H.~Zhang, D.~Jin, A.~Jasour, M.~Ghaffari, and L.~Carlone, ``Gmkf: Generalized moment kalman filter for polynomial systems with arbitrary noise,'' \emph{arXiv preprint arXiv:2403.04712}, 2024.

\bibitem{jin2024multi}
D.~Jin, S.~Karmalkar, H.~Zhang, and L.~Carlone, ``Multi-model 3d registration: Finding multiple moving objects in cluttered point clouds,'' \emph{arXiv preprint arXiv:2402.10865}, 2024.

\bibitem{yao2023apla}
Y.~Yao, S.~Deng, Z.~Cao, H.~Zhang, and L.-J. Deng, ``Apla: Additional perturbation for latent noise with adversarial training enables consistency,'' \emph{arXiv preprint arXiv:2308.12605}, 2023.

\bibitem{shen2024diffclip}
S.~Shen, Z.~Zhu, L.~Fan, H.~Zhang, and X.~Wu, ``Diffclip: Leveraging stable diffusion for language grounded 3d classification,'' in \emph{Proceedings of the IEEE/CVF Winter Conference on Applications of Computer Vision}, 2024, pp. 3596--3605.

\bibitem{lim2021planar}
V.~Lim, H.~Huang, L.~Y. Chen, J.~Wang, J.~Ichnowski, D.~Seita, M.~Laskey, and K.~Goldberg, ``Planar robot casting with real2sim2real self-supervised learning,'' \emph{arXiv preprint arXiv:2111.04814}, 2021.

\bibitem{lim2022real2sim2real}
------, ``Real2sim2real: Self-supervised learning of physical single-step dynamic actions for planar robot casting,'' in \emph{2022 International Conference on Robotics and Automation (ICRA)}.\hskip 1em plus 0.5em minus 0.4em\relax IEEE, 2022, pp. 8282--8289.

\bibitem{cs184}
C.~184, ``Assignment 4: Cloth simulation,'' 2019.

\bibitem{schmidt2016self}
T.~Schmidt, R.~Newcombe, and D.~Fox, ``Self-supervised visual descriptor learning for dense correspondence,'' \emph{IEEE Robotics and Automation Letters}, vol.~2, no.~2, pp. 420--427, 2016.

\bibitem{lowe2004sift}
D.~Lowe, ``Distinctive image features from scale-invariant keypoints,'' in \emph{IJCV}, 2004, pp. 91--110.

\bibitem{bay2008surf}
H.~Bay, T.~Tuytelaars, and L.~V. Gool, ``Speeded-up robust features (surf),'' in \emph{ECCV}, 2006, pp. 404--417.

\bibitem{rublee2011orb}
E.~Rublee, V.~Rabaud, K.~Konolige, and G.~R. Bradski, ``Orb: An efficient alternative to sift or surf.'' in \emph{ICCV}, vol.~11, no.~1.\hskip 1em plus 0.5em minus 0.4em\relax Citeseer, 2011, p.~2.

\bibitem{wang2019learning}
X.~Wang, A.~Jabri, and A.~A. Efros, ``Learning correspondence from the cycle-consistency of time,'' in \emph{Proceedings of the IEEE Conference on Computer Vision and Pattern Recognition}, 2019, pp. 2566--2576.

\bibitem{dense-obj-nets}
P.~R. Florence, L.~Manuelli, and R.~Tedrake, ``Dense object nets: Learning dense visual object descriptors by and for robotic manipulation,'' in \emph{CoRL}, 2018.

\end{thebibliography}

\appendix

\end{document}